\documentclass[10pt,twocolumn,letterpaper]{article}

\usepackage[pagenumbers]{cvpr} 

\usepackage[accsupp]{axessibility}
\usepackage{algorithm}
\usepackage{algpseudocodex}
\usepackage{amsmath}
\usepackage{amssymb}
\usepackage{multirow}
\usepackage{booktabs}
\usepackage{makecell}
\usepackage{subcaption}
\usepackage{fancybox}
\usepackage{fancyvrb}
\usepackage{bm}
\usepackage{multirow}
\usepackage{booktabs}
\usepackage{bbding}
\usepackage{makecell}
\usepackage{subcaption}
\usepackage{fancybox}
\usepackage{fancyvrb}

\definecolor{cvprblue}{rgb}{0.21,0.49,0.74}
\usepackage[pagebackref,breaklinks,colorlinks,citecolor=cvprblue]{hyperref}

\title{Gen-LangSplat: Generalized Language Gaussian Splatting with Pre-Trained Feature Compression}

\author{
  Pranav Saxena$^{1}$
}

\begin{document}

\twocolumn[{%
\renewcommand\twocolumn[1][]{#1}%
\maketitle
\begin{center}
    \centering
    \captionsetup{type=figure}
    \includegraphics[width=1\textwidth]{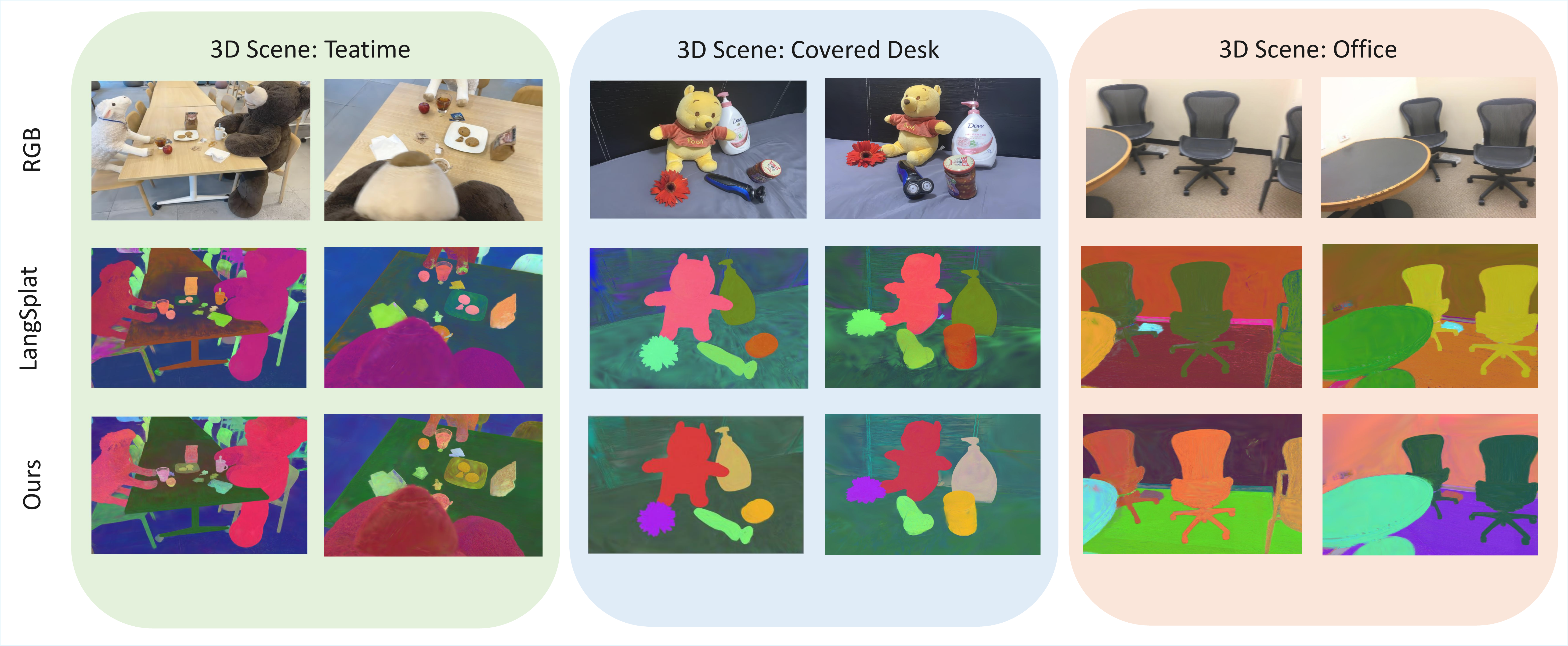}
    \captionof{figure}{Visualization of learned 3D language features of the previous SOTA method, LangSplat, and our proposed approach. Without requiring per-scene training for the language-feature autoencoder, our method achieves comparable, or even superior, results while being more efficient.
    }
    \label{fig:teaser}
\vspace{-6pt}
\end{center}%
}]
{\let\thefootnote\relax\footnotetext{$^{1}$Pranav Saxena is with Birla Institute of
Technology and Science Pilani, K.K Birla Goa Campus, Goa, India. This work was done in collaboration with National University of Singapore and TU Delft, Netherlands.}

\begin{abstract}
Modeling open-vocabulary language fields in 3D is essential for intuitive human-AI interaction and querying within physical environments. State-of-the-art approaches, such as LangSplat, leverage 3D Gaussian Splatting to efficiently construct these language fields, encoding features distilled from high-dimensional models like CLIP. However, this efficiency is currently offset by the requirement to train a scene-specific language autoencoder for feature compression, introducing a costly, per-scene optimization bottleneck that hinders deployment scalability.
In this work, we introduce Gen-LangSplat, that eliminates this requirement by replacing the scene-wise autoencoder with a generalized autoencoder, pre-trained extensively on the large-scale ScanNet dataset. This architectural shift enables the use of a fixed, compact latent space for language features across any new scene without any scene-specific training. By removing this dependency, our entire language field construction process achieves a  efficiency boost while delivering querying performance comparable to, or exceeding, the original LangSplat method. To validate our design choice, we perform a thorough ablation study empirically determining the optimal latent embedding dimension and quantifying representational fidelity using Mean Squared Error and cosine similarity between the original and reprojected 512-dimensional CLIP embeddings. Our results demonstrate that generalized embeddings can efficiently and accurately support open-vocabulary querying in novel 3D scenes, paving the way for scalable, real-time interactive 3D AI applications. The code can be found at \url{https://github.com/Pranav-Saxena/Gen-LangSplat}.
\end{abstract}

\section{Introduction}
\label{sec:intro}

Understanding and interacting with the 3D world through natural language is an emerging challenge at the intersection of 3D vision and multimodal learning. The ability to associate linguistic concepts with spatially grounded representations is essential for a range of applications in embodied AI, including open-vocabulary scene understanding, 3D semantic reasoning, and language-guided robotic manipulation. Earlier methods have primarily relied on Neural Radiance Fields (NeRFs) that distill vision-language model (VLM) features, such as those from CLIP~\cite{radford2021learning}, into volumetric neural fields. While these approaches enable open-ended language queries within reconstructed 3D scenes, the implicit nature of NeRFs introduces substantial inefficiencies, rendering is slow, optimization requires dense sampling, and the models are difficult to adapt or edit. These limitations have motivated the exploration of more explicit and computationally efficient 3D representations.

Recent advances in 3D Gaussian Splatting (3DGS)~\cite{kerbl20233d} have demonstrated remarkable efficiency and visual fidelity by representing scenes as a collection of spatially distributed anisotropic Gaussians. Building on this representation, LangSplat~\cite{qin2023langsplat} introduced the concept of a 3D language field, where CLIP features are attached to individual Gaussians to support open-vocabulary querying in 3D. By replacing volumetric NeRF rendering with tile-based Gaussian rasterization, LangSplat achieves over two orders of magnitude faster performance while maintaining high-quality alignment between image and language spaces. However, a key limitation of LangSplat lies in its reliance on per-scene autoencoders, which are trained separately for every scene to compress high-dimensional CLIP features into a low-dimensional latent space. This per-scene feature compression not only adds significant computational overhead but also prevents generalization to unseen scenes, since each environment requires new optimization to adapt the language embeddings.

In this work, we revisit this design bottleneck and propose a framework that eliminates the need for scene-specific autoencoders through the introduction of a generalized autoencoder trained across diverse scenes. Our method leverages a large-scale dataset, ScanNet~\cite{dai2017scannet}, to train a unified encoder--decoder network capable of learning a compact and transferable latent space for CLIP embeddings. Once trained, the generalized autoencoder can directly process CLIP features from new scenes without any fine-tuning, enabling rapid deployment across diverse environments. Integrating this generalized feature compression into a 3D Gaussian Splatting pipeline preserves the open-vocabulary grounding ability of LangSplat while improving overall efficiency by approximately $2\times$. Each Gaussian in our model jointly encodes appearance, geometry, and low-dimensional language features derived from the generalized embedding, which are decoded back to CLIP space during rendering to ensure multi-view consistency. This design not only accelerates training and inference but also establishes a consistent latent manifold shared across different scenes, facilitating scalable and cross-scene language reasoning in 3D.

To evaluate the representational strength of our generalized embedding, we perform a comprehensive ablation study that investigates both efficiency and reconstruction fidelity. We vary the latent dimensionality to analyze its effect on rendering quality, memory cost, and retrieval accuracy, and further quantify how closely reprojected CLIP embeddings approximate their original 512-dimensional counterparts using mean squared error (MSE) and cosine similarity metrics. Our findings reveal that a 16-dimensional embedding achieves the optimal balance between compactness and fidelity, outperforming the 3-dimensional latent space used in LangSplat in both reconstruction accuracy and semantic consistency. Despite removing the per-scene training step, our method achieves comparable localization precision and open-vocabulary retrieval accuracy to LangSplat, confirming that a single generalized autoencoder is sufficient for language-grounded 3D reconstruction across diverse real-world environments.

In summary, our work establishes that cross-scene feature compression is a viable and efficient alternative to scene-specific optimization, making language-aware 3D Gaussian Splatting more practical and effective. By decoupling the language embedding process from per-scene adaptation, we demonstrate that robust open-vocabulary 3D understanding can be achieved at a fraction of the cost and time.
We summarize the contributions of this paper as follows:
\begin{itemize}
    \item We introduce a cross-scene generalized autoencoder trained on the ScanNet dataset that learns a transferable latent representation for language features. Unlike prior per-scene approaches, our model eliminates the need for scene-specific retraining while maintaining strong semantic consistency across diverse environments.
    
    \item Our method achieves nearly a 2× improvement in overall efficiency compared to LangSplat by removing the need to train a separate autoencoder for each scene.
    
    \item We conduct a detailed ablation study on latent dimensionality to analyze its impact on feature reconstruction and semantic retention. The results show that a 16-dimensional latent embedding achieves high cosine similarity and low MSE when reprojected into the CLIP feature space, without compromising efficiency.
\end{itemize}

\section{Related Work}
\label{sec:related}

\subsection{3D Scene Representations and Neural Rendering}
Neural implicit representations have been widely adopted for high-quality 3D reconstruction and rendering. Early works such as DeepVoxels~\cite{sitzmann2019deepvoxels} and Neural Volumes~\cite{Lombardi2019} introduced volumetric neural rendering pipelines that model scene appearance as continuous functions. Neural Radiance Fields (NeRFs)~\cite{mildenhall2020nerf} further revolutionized this field by representing radiance and density using multilayer perceptrons, achieving photorealistic view synthesis from sparse inputs. However, the implicit nature of NeRFs incurs heavy computational cost due to dense ray sampling and global optimization, limiting their real-time applicability. Later extensions such as Mip-NeRF~\cite{barron2021mipnerf}, Plenoxels~\cite{yu_and_fridovichkeil2021plenoxels}, and Instant-NGP~\cite{muller2022instant} improved speed and efficiency using mip-level sampling, voxel grids, and hash-based encodings, yet still rely on costly volume integration.  

3D Gaussian Splatting (3DGS)~\cite{kerbl20233d} introduced a paradigm shift by representing scenes as explicit, spatially adaptive anisotropic Gaussians optimized through differentiable rasterization. Each Gaussian encapsulates geometry, color, opacity, and orientation, enabling real-time rendering with significantly reduced memory usage. Subsequent works have demonstrated the flexibility of this representation for applications in novel-view synthesis~\cite{cheng2024gaussianpro}, dynamic scenes~\cite{Yang2023FreeNeRF}, and efficient reconstruction from sparse observations~\cite{xiong2023sparsegs}. This explicit and differentiable formulation serves as the foundation for recent developments in semantically and linguistically grounded 3D understanding.

\subsection{Language-Grounded 3D Representations}
Integrating vision--language understanding into 3D scene representations has become increasingly relevant for open-vocabulary reasoning and interactive scene understanding. The success of large-scale vision--language models such as CLIP~\cite{radford2021learning} has inspired a series of methods that incorporate language embeddings into neural rendering frameworks. Early works like LERF~\cite{kerr2023lerf} and Semantic-NeRF~\cite{zhi2021place} embedded CLIP features within NeRF fields to enable open-ended text queries and semantic localization in 3D. Despite their strong alignment between vision and language modalities, these methods inherit NeRF’s inefficiencies in rendering and optimization, constraining their scalability to large or complex scenes.

LangSplat~\cite{qin2023langsplat} overcomes these limitations by introducing a 3D Gaussian-based language field that attaches CLIP features to individual Gaussians, achieving precise and efficient open-vocabulary 3D querying. Through hierarchical SAM-based supervision~\cite{kirillov2023segment}, LangSplat delineates fine-grained object boundaries and reduces feature bleeding across regions. However, its reliance on per-scene autoencoders for compressing 512-dimensional CLIP embeddings into a compact latent space significantly restricts scalability. Each scene requires a new autoencoder to be trained, resulting in additional compute overhead and lack of cross-scene generalization.  

Our approach eliminates this bottleneck by introducing a generalized language autoencoder trained on ScanNet~\cite{dai2017scannet}, which directly generalizes across unseen environments without retraining.

\subsection{3D Grouping and Semantic Decomposition}
Recent works have explored enhancing interpretability and compositional reasoning within Gaussian-based frameworks. Gaussian Grouping (GG)~\cite{gaussian_grouping} introduces learnable identity embeddings per Gaussian, enabling zero-shot instance segmentation and 3D object grouping. Similar methods such as EditSplat~\cite{lee2024editsplat} and SAGS~\cite{hu2024semantic} leverage SAM~\cite{kirillov2023segment} or DINO~\cite{caron2021emerging} features to guide 3D semantic decomposition, allowing downstream applications such as object editing, removal, and part-level manipulation. While these methods enhance geometric interpretability, they do not incorporate language grounding and thus cannot support open-vocabulary querying. Our framework complements these efforts by coupling semantic reasoning with efficient, language-aware 3D Gaussian representations.

\section{Preliminaries}
\label{sec:prelim}

\begin{figure*}[h!]
    \centering
    \includegraphics[width=0.98\textwidth]{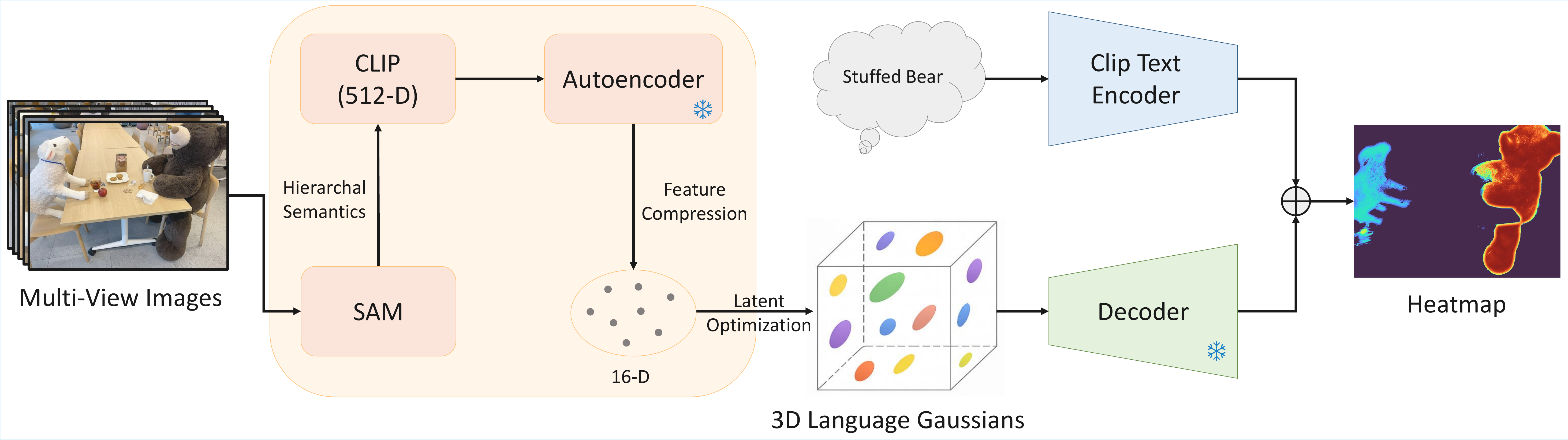}
    \caption{Overview of the proposed Gen-LangSplat framework. We leverage SAM to extract hierarchical semantics from multi-view images to resolve point ambiguity. The resulting segmentation masks are processed by the CLIP image encoder to obtain 512-D embeddings. These embeddings are compressed into a 16-D latent space using a generalized autoencoder pre-trained on ScanNet. Our 3D language Gaussians learn language features directly on a shared latent space derived from the pre-trained autoencoder. During querying, the rendered latent embeddings are decoded through the frozen decoder to recover the corresponding CLIP-space features for semantic reasoning.
    }  \label{fig:framework}
    \vspace{-14pt}
\end{figure*}

\noindent \textbf{3D Gaussian Representation.}
A 3D scene can be represented explicitly as a set of anisotropic Gaussians distributed in space, where each Gaussian $G_i(x)$ is defined by a mean $\mu_i \in \mathbb{R}^3$ and a covariance matrix $\Sigma_i \in \mathbb{R}^{3 \times 3}$:
\begin{equation}
G_i(x) = \exp\left(-\frac{1}{2}(x - \mu_i)^\top \Sigma_i^{-1} (x - \mu_i)\right).
\end{equation}
Each Gaussian additionally stores color $\mathbf{c}_i \in \mathbb{R}^3$, opacity $\alpha_i$, and a scale-rotation decomposition of $\Sigma_i$ to model spatial extent and orientation. The full scene is thus represented as $\mathcal{G} = \{(\mu_i, \Sigma_i, \mathbf{c}_i, \alpha_i)\}_{i=1}^N$ for $N$ Gaussians.

\vspace{0.5\baselineskip}
\noindent \textbf{Differentiable Rendering.}
Rendering proceeds by projecting each 3D Gaussian into the image plane using a perspective projection $\pi(\cdot)$ parameterized by camera intrinsics and extrinsics. The projected 2D Gaussian contributes to pixel $v$ according to its screen-space extent and accumulated opacity. The rendered color at pixel $v$ is obtained through front-to-back alpha compositing:
\begin{equation}
C(v) = \sum_{i \in \mathcal{N}(v)} \mathbf{c}_i \, \alpha_i \prod_{j=1}^{i-1}(1 - \alpha_j),
\end{equation}
where $\mathcal{N}(v)$ denotes the ordered set of Gaussians intersecting the camera ray corresponding to pixel $v$, $c_i$ is the color of the $i$-th Gaussian, and $\alpha_i = o_i G^{2D}_i(v)$. Here $o_i$ is the opacity of the $i$-th Gaussian and $G^{2D}_i(\cdot)$ represents the function of the $i$-th Gaussian projected onto 2D.. This formulation provides a fully differentiable rasterization pipeline that allows gradients to            from image-space loss functions to the 3D Gaussian parameters.

\vspace{0.5\baselineskip}
\noindent \textbf{Language Feature Embedding.}
To enable semantic reasoning and open-vocabulary understanding, each Gaussian is augmented with an additional feature vector $\mathbf{f}_i \in \mathbb{R}^d$ derived from a vision--language model such as CLIP. These feature vectors encode language-aligned semantics corresponding to the local image region or object represented by the Gaussian. During rendering, the per-pixel language feature is accumulated analogously to color:
\begin{equation}
F(v) = \sum_{i \in \mathcal{N}(v)} \mathbf{f}_i \, \alpha_i \prod_{j=1}^{i-1}(1 - \alpha_j).
\end{equation}
The resulting feature map $F(v)$ can then be used for open-vocabulary querying or similarity-based retrieval in the language embedding space.  

\vspace{0.5\baselineskip}
\noindent \textbf{Feature Compression.}
Since high-dimensional embeddings (e.g., 512-D CLIP features) are expensive to store and optimize for every Gaussian, a learned encoder--decoder network is employed to compress the language features into a low-dimensional latent space $\mathbf{z}_i \in \mathbb{R}^k$, where $k \ll 512$. The encoder $E(\cdot)$ and decoder $D(\cdot)$ satisfy
\begin{equation}
\mathbf{z}_i = E(\mathbf{f}_i), \qquad \hat{\mathbf{f}}_i = D(\mathbf{z}_i),
\end{equation}
and are trained to minimize reconstruction loss between $\hat{\mathbf{f}}_i$ and $\mathbf{f}_i$:
\begin{equation}
\mathcal{L}_{\text{rec}} = \|\hat{\mathbf{f}}_i - \mathbf{f}_i\|_2^2 + \lambda (1 - \cos(\hat{\mathbf{f}}_i, \mathbf{f}_i)).
\end{equation}
This compression preserves semantic consistency while significantly reducing memory and computation costs, enabling scalable, language-aware 3D Gaussian rendering.

\section{Methodology}
\label{sec:method}

Our method builds upon the LangSplat framework and extends it with a generalized language feature compression module to enable scalable, memory-efficient open-vocabulary 3D reasoning. The overall pipeline is composed of three main stages: \textit{(1) Preprocessing}, where hierarchical semantics and CLIP features are prepared; \textit{(2) Generalized Autoencoder}, where a pre-trained cross-scene encoder compresses CLIP embeddings into a 16-dimensional latent space; and \textit{(3) Training of Language-Embedded Gaussians}, where the pre-trained RGB Gaussians are taken as initialization, and only the language feature channels are optimized to learn the 3D language field within the Gaussian Splatting framework.

\subsection{Preprocessing}

Given a set of posed RGB images $\{I_t\}_{t=1}^{T}$, we begin by extracting hierarchical semantic information to mitigate the point ambiguity problem, where a single 3D point contributes to multiple overlapping semantic levels. To this end, we leverage the Segment Anything Model (SAM) to produce multi-scale hierarchical segmentation masks for each image. These masks serve as spatially coherent supervision regions that guide the extraction of semantically meaningful features.

Each segmented region is passed through the CLIP image encoder, resulting in a 512-dimensional embedding $\mathbf{f}_m \in \mathbb{R}^{512}$ for each mask $m$. These embeddings capture the high-level semantic correspondence between visual regions and textual concepts. Compared to patch-based multi-scale methods that suffer from feature smoothing and boundary leakage, the use of SAM-based masks ensures sharper object boundaries and semantically aligned regions across multiple levels of granularity. The resulting set of CLIP embeddings forms the high-dimensional input feature space that will later be compressed using our generalized autoencoder.

\subsection{Generalized Autoencoder}

Instead of training a per-scene language autoencoder as in LangSplat, we propose a single generalized autoencoder trained across diverse indoor environments to learn a unified and transferable latent representation for CLIP features. The network consists of an encoder $E_\theta$ that maps high-dimensional CLIP embeddings to a compact latent space, and a decoder $D_\phi$ that reconstructs them back to the original CLIP feature space:
\begin{equation}
\mathbf{z} = E_\theta(\mathbf{f}), \quad \hat{\mathbf{f}} = D_\phi(\mathbf{z}),
\end{equation}
where $\mathbf{f} \in \mathbb{R}^{512}$ denotes the CLIP feature and $\mathbf{z} \in \mathbb{R}^{k}$ is the latent representation. Following LangSplat, the network is optimized using an $\ell_1$ reconstruction loss combined with a cosine similarity term:
\begin{equation}
\mathcal{L}_{AE} = \|\mathbf{f} - \hat{\mathbf{f}}\|_1 + \lambda(1 - \cos(\mathbf{f}, \hat{\mathbf{f}})),
\end{equation}
ensuring both feature reconstruction accuracy and angular consistency within CLIP space.

The autoencoder is pre-trained on CLIP embeddings extracted from SAM-derived segmentation masks over ScanNet scenes. For each image, SAM (ViT-H) generates dense mask proposals, and OpenCLIP ViT-B/16 encodes these masked regions into 512-dimensional features. Training across millions of such mask-level embeddings enables the autoencoder to capture category and scene-invariant semantic structure. This cross-scene supervision allows direct feature compression at test time without requiring any scene-specific optimization.

Through ablation, we find that a latent dimensionality of $k=16$ retains over 93\% cosine similarity to the original CLIP embeddings, while providing a significant reduction in storage and computational cost. After pre-training, both the encoder and decoder weights are frozen for all downstream tasks, allowing immediate deployment on novel scenes with no additional fine-tuning.

\subsection{Training of Language-Embedded Gaussians}

We integrate the generalized autoencoder into the 3D Gaussian Splatting framework to construct a compact 3D language field. Each Gaussian $G_i = (\mu_i, \Sigma_i, \mathbf{c}_i, \alpha_i, \mathbf{z}_i)$ represents spatial position $\mu_i$, covariance $\Sigma_i$, color coefficients $\mathbf{c}_i$, opacity $\alpha_i$, and a learnable latent language feature $\mathbf{z}_i$ encoded by $E_\theta$. 

Following the standard 3DGS pipeline, we first optimize all Gaussians for photometric reconstruction using RGB supervision. Once the RGB model converges, all geometric and appearance parameters ($\mu_i$, $\Sigma_i$, $\mathbf{c}_i$, $\alpha_i$) are frozen, and only the latent language features $\mathbf{z}_i$ are optimized. For each camera view, the rendered latent feature map is obtained by alpha compositing the latent codes:
\begin{equation}
Z(v) = \sum_{i \in \mathcal{N}(v)} \mathbf{z}_i \, \alpha_i(v) \prod_{j < i} (1 - \alpha_j(v)),
\end{equation}
where $\mathcal{N}(v)$ denotes the ordered set of Gaussians along the ray corresponding to pixel $v$, and $\alpha_i(v)$ is the view-dependent opacity. The target features are computed by encoding per-mask CLIP features of the input image using the frozen encoder $E_\theta$. The rendered and target latent maps are supervised using a combined $\ell_1$ and cosine similarity loss:
\begin{equation}
\mathcal{L}_{Lang} = \|Z(v) - H(v)\|_1 + \gamma(1 - \cos(Z(v), H(v))).
\end{equation}
The overall training objective is then:
\begin{equation}
\mathcal{L}_{total} = \mathcal{L}_{RGB} + \beta \, \mathcal{L}_{Lang},
\end{equation}
where $\mathcal{L}_{RGB}$ enforces photometric consistency from the RGB stage, and $\beta$ balances semantic supervision.

Unlike prior methods that train a separate autoencoder per scene, our design enables the 3D Gaussian language field to operate directly on a shared latent manifold. This leads to a two-fold reduction in total training time, consistent feature distributions across scenes, and comparable open-vocabulary localization performance to scene-specific approaches. During querying, the rendered latent embeddings are decoded via $D_\phi$ into CLIP space, allowing direct text-based retrieval and open-vocabulary interaction.

\section{Experiments and Results}
\label{sec:experiment}

\subsection{Evaluation}

\begin{figure*}[t]
  \centering
   \includegraphics[width=1\linewidth]{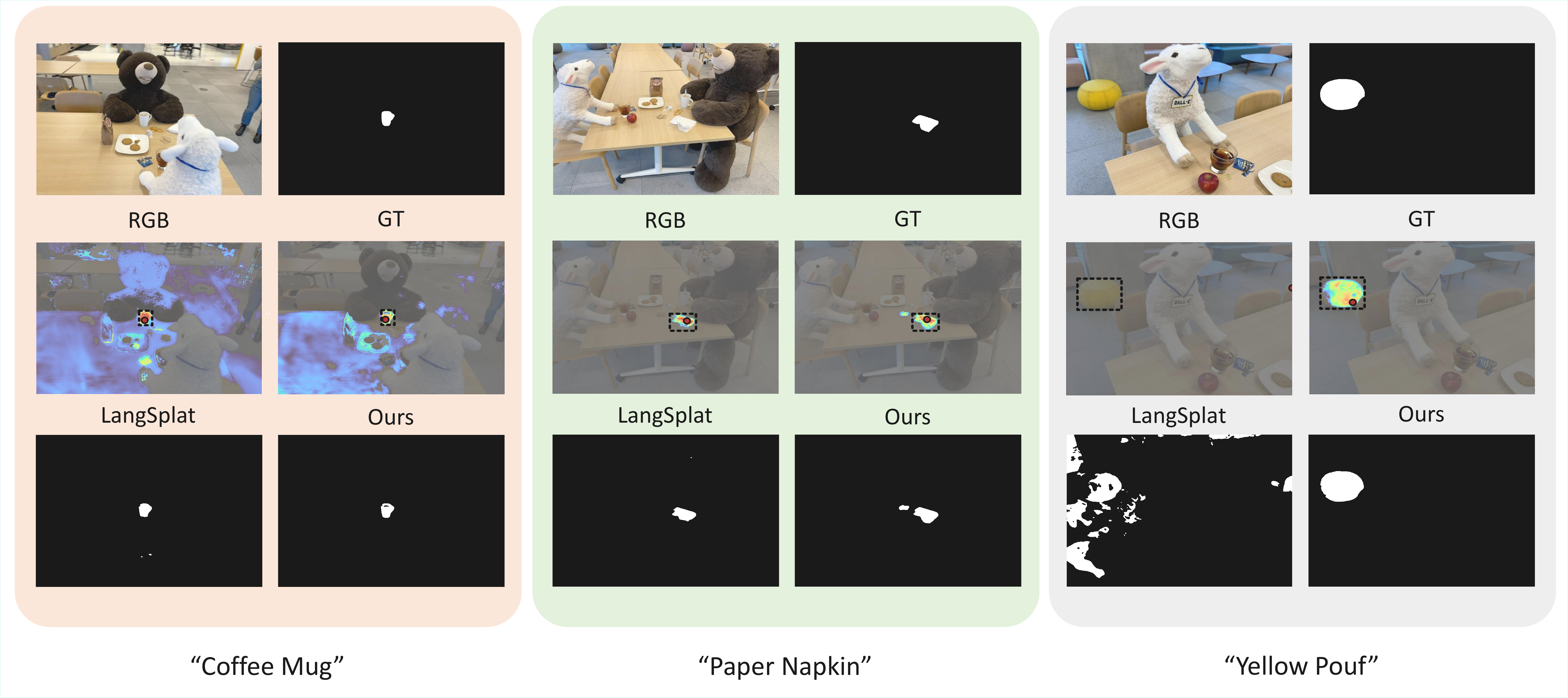}
   \caption{Qualitative comparison of open-vocabulary 3D object localization and segmentation results on the LERF dataset. The red points indicate model predictions, black dashed boxes denote the ground-truth annotations, and the bottom row shows the corresponding binary segmentation masks.}
   \label{fig:localization}
    \vspace{-4pt}
\end{figure*}

\textbf{Datasets.} Similar to LangSplat, we evaluate our approach on two datasets.
(i) LERF dataset~\cite{kerr2023lerf}, which consists of complex in-the-wild scenes designed for 3D object localization tasks. We use the extended version of LERF provided by LangSplat, which includes annotated ground-truth masks for textual queries, enabling open-vocabulary 3D semantic segmentation evaluation. Following LERF~\cite{kerr2023lerf}, we report localization accuracy for the 3D object localization task and Intersection-over-Union (IoU) for 3D semantic segmentation.
(ii) 3D-OVS dataset~\cite{liu2023weakly}, which contains a diverse collection of long-tail indoor objects captured under varying poses and backgrounds. This dataset is designed for open-vocabulary 3D semantic segmentation, where the full list of object categories is available. Unlike other methods that use the entire list to generate predicted masks, we use only the queried category to produce the corresponding masks. The mean Intersection-over-Union (mIoU) metric is used for evaluation on this dataset.

\noindent\textbf{Implementation Details.}
To extract the language features for each image, we employ the OpenCLIP ViT-B/16 model. For segmentation, we use the Segment Anything Model (SAM) with the ViT-H backbone to generate hierarchical 2D masks. For each scene, we first train a 3D Gaussian Splatting (3DGS) model to reconstruct the RGB scene following the default hyperparameter settings in~\cite{kerbl20233d}. The model is trained for 30,000 iterations, resulting in approximately 2.5 million Gaussians per scene.
Then we train our 3D language Gaussians by fixing all other parameters of 3D Gaussians such as mean and opacity. Only the language features are learnable during this stage. We train the language features for 30,000 iterations.
Our generalized autoencoder is implemented as an MLP that compresses 512-dimensional CLIP embeddings into 16-dimensional latent features. The autoencoder is trained on the ScanNet dataset using an NVIDIA RTX 4090 GPU (24 GB VRAM).


\subsection{Results on the LERF dataset}

\textbf{Quantitative Results.} We first evaluate our method on the LERF dataset to compare its performance against existing approaches. The localization results are presented in Table~\ref{table:lerf_loca}. Our method achieves comparable, and in some cases superior, performance to LangSplat, attaining an overall localization accuracy of 84.4\%. Table~\ref{table:lerf_seg} reports the IoU scores for 3D semantic segmentation, where our approach performs on par with LangSplat. These results demonstrate the effectiveness of our generalized autoencoder design, confirming that a single cross-scene feature compressor can replace scene-specific training without degrading performance.

\begin{table}[H]
  \centering
  \begin{tabular}{lccc}
    \toprule
    Test Scene &  LERF~\cite{kerr2023lerf} & LangSplat~\cite{qin2023langsplat} & Ours \\
    \midrule
    ramen & 62.0 & \textbf{73.2} &  72.8 \\
    figurines &  75.0 & 80.4 & \textbf{80.7} \\
    teatime  & 84.8 & 88.1 & \textbf{88.9} \\
    waldo\_kitchen  & 72.7 & \textbf{95.5} & 95.1 \\
    \midrule
    overall  & 73.6 & 84.3 & \textbf{84.4} \\
    \bottomrule
  \end{tabular}
  \caption{Localization accuracy (\%) comparisons on LERF dataset.}
  \label{table:lerf_loca}
  \vspace{-5pt}
\end{table}

\begin{figure*}[t]
  \centering
   \includegraphics[width=1\linewidth]{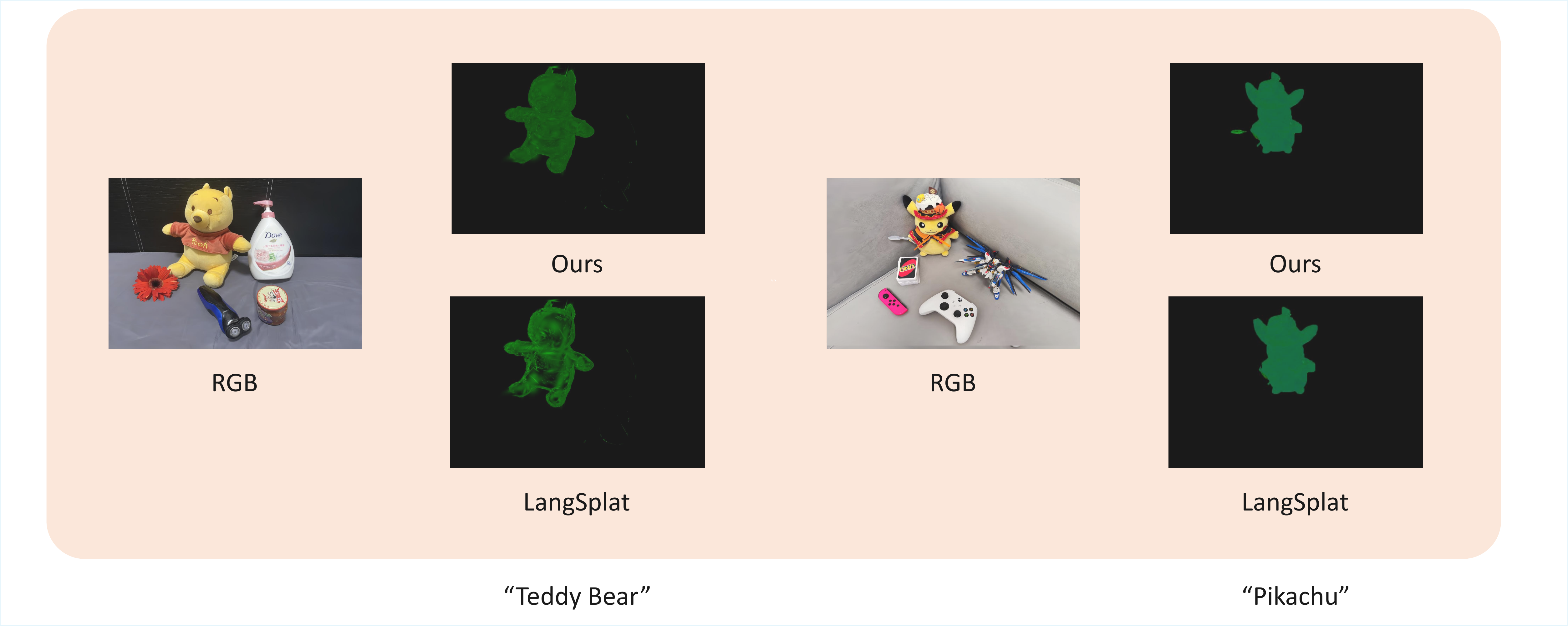}
   \caption{Qualitative comparisons of 3D Segmentation on the 3D OVS Dataset.}
   \label{fig:3dovs}
   \vspace{-14pt}
\end{figure*}

\noindent\textbf{Visualization Results.} To visualize the learned 3D language field, we project the learned language features onto their principal components using PCA, following previous work~\cite{kobayashi2022decomposing}. The visualization results are shown in Figure~\ref{fig:teaser}. We observe that the features learned by our method exhibit similar, and in some cases superior, structure compared to LangSplat, particularly in the Teatime scene. Furthermore, we present qualitative results for 3D object localization and semantic segmentation in Figure~\ref{fig:localization}.

\begin{table}[H]
  \centering
  \begin{tabular}{lccc}
    \toprule
    Test Scene & LERF~\cite{kerr2023lerf} & LangSplat~\cite{qin2023langsplat} & Ours \\
    \midrule
    ramen &  28.2 &  \textbf{51.2} & 50.9 \\
    figurines & 38.6 & 44.7 & \textbf{45.3} \\
    teatime  & 45.0 & 65.1  & \textbf{65.8} \\
    waldo\_kitchen & 37.9 & \textbf{44.5} & 44.3 \\
    \midrule
    overall & 37.4 & 51.4 & \textbf{51.6} \\
    \bottomrule
  \end{tabular}
  \caption{Quantitative comparisons of 3D semantic segmentation on the LERF dataset. We report the average IoU scores (\%).}
  \label{table:lerf_seg}
  \vspace{-5pt}
\end{table}


\subsection{Results on the 3D-OVS dataset} 

\textbf{Quantitative Results.} We compare our method with other 2D and 3D state-of-the-art approaches on the 3D-OVS dataset, as shown in Table~\ref{table:3dovs}. Our method achieves performance comparable to or exceeding existing approaches. Specifically, it performs on par with LangSplat while surpassing 2D-based methods such as ODISE~\cite{xu2023odise} and OV-Seg~\cite{liang2023open}, as well as 3D-based methods including LERF~\cite{kerr2023lerf} and 3D-OVS~\cite{liu2023segment}. Notably, for this dataset, we generate object masks using only the query category, while other methods (e.g., 3D-OVS) require the complete list of categories for prediction. Overall, our approach attains an mIoU of 93.3\%, demonstrating that the proposed generalized autoencoder effectively retains semantic consistency and enables accurate open-vocabulary 3D segmentation without per-scene training.

\begin{table}[H]
  \renewcommand\tabcolsep{4pt}
  \centering
  \begin{tabular}{lcccccc}
    \toprule
    Method & \textit{bed}   & \textit{bench} & \textit{room}  & \textit{sofa}  & \textit{lawn}  & overall   \\
    \midrule
    LSeg~\cite{li2022languagedriven}      & 56.0           & 6.0            & 19.2           & 4.5            & 17.5           & 20.6           \\
    ODISE~\cite{xu2023odise}     & 52.6           & 24.1           & 52.5           & 48.3           & 39.8           & 43.5           \\
    OV-Seg~\cite{liang2023open}    & 79.8           & 88.9           & 71.4           & 66.1           & 81.2           & 77.5           \\
    \midrule
    FFD~\cite{kobayashi2022decomposing}       & 56.6           & 6.1            & 25.1           & 3.7            & 42.9           & 26.9           \\
    LERF~\cite{kerr2023lerf}      & 73.5           & 53.2           & 46.6           & 27             & 73.7           & 54.8            \\
    3D-OVS~\cite{liu2023weakly}    & 89.5           & 89.3           & 92.8           & 74             & 88.2           & 86.8           \\
    LangSplat~\cite{qin2023langsplat}   & \textbf{92.5} & 94.2 & 94.1 & \textbf{90.0} & \textbf{96.1} & \textbf{93.4} \\ 
    \midrule
    Ours   & 92.1 & \textbf{94.6} & \textbf{94.4} & 89.4 & 95.9 & 93.3 \\ 
    \bottomrule
  \end{tabular}
  \caption{Quantitative comparisons of 3D semantic segmentation on the 3D-OVS dataset. We report the mIoU scores (\%).}
  \label{table:3dovs}
\end{table}

\textbf{Open Vocabulary Segmentation.} We perform open vocabulary segmentation on the 3D-OVS dataset given a language query. We present the results in Figure~\ref{fig:3dovs}. Cosine similarity is used to evaluate correspondence between the rendered 3D language embeddings and the query text, used to retrieve gaussians with the highest semantic alignment.

\subsection{Ablation Study}

To analyze how effectively different latent dimensions retain semantic information from CLIP embeddings, we conduct an ablation study by varying the latent space dimension \( d \). For each configuration, we evaluate the reconstructed CLIP features obtained from the decoder by computing the mean squared error (MSE) and cosine similarity with respect to the original 512-dimensional embeddings. The results are shown in Figure~\ref{fig:clip_ablation}.  

\begin{figure}[H]
    \centering
    \includegraphics[width=1.0\linewidth]{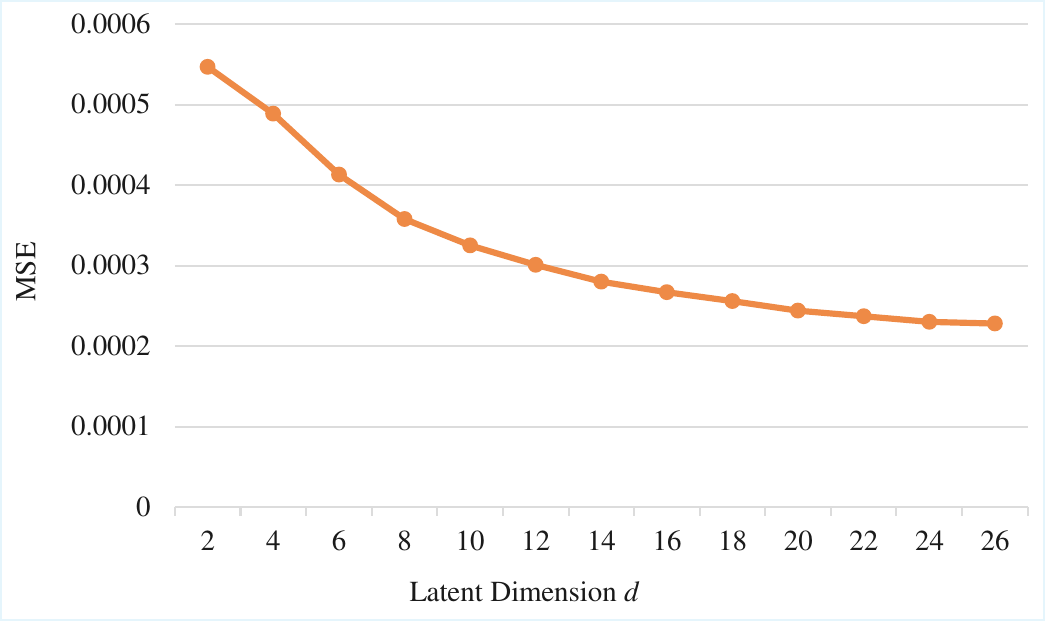}
    \vspace{3pt} 
    \includegraphics[width=1.0\linewidth]{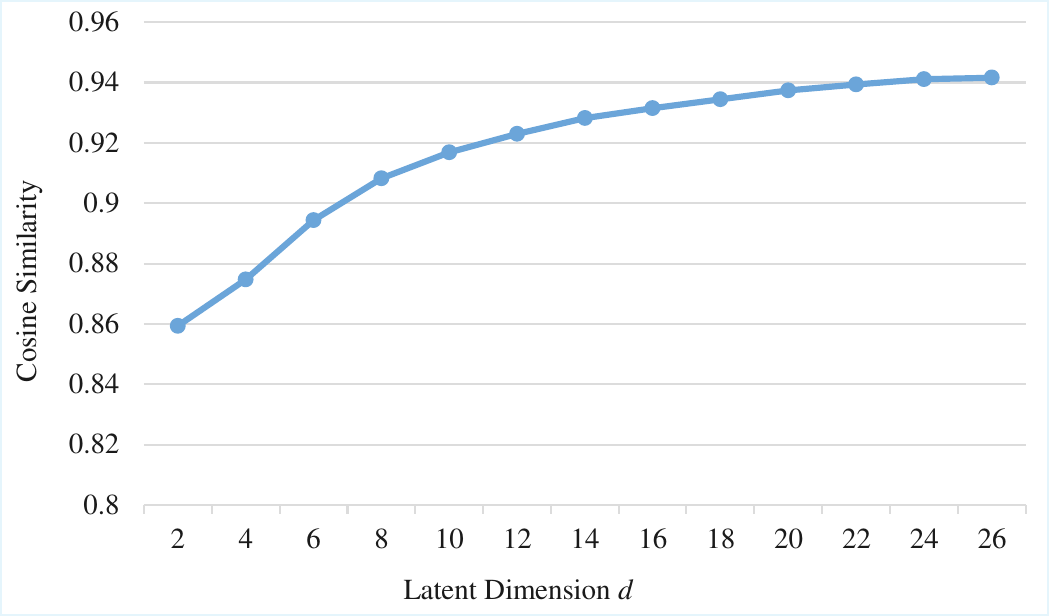}
    \caption{Ablation study on the latent dimensionality of our generalized autoencoder. As the dimension increases, the reconstruction error (MSE) decreases while the cosine similarity improves and saturates beyond \(d=16\), 
    indicating an optimal trade-off between compactness and semantic retention.}
    \label{fig:clip_ablation}
\end{figure}

This study aims to determine which latent dimension can best preserve the original CLIP feature information after compression and reprojection. As observed, increasing \( d \) consistently improves reconstruction quality, MSE decreases while cosine similarity increases and begins to saturate beyond \( d = 16 \). At this point, the MSE reaches approximately \( 3 \times 10^{-4} \) and cosine similarity exceeds 0.93, indicating that most semantic information from the original features is retained.

These findings suggest that a 16-dimensional latent space provides the best trade-off between representational fidelity and compactness, effectively preserving language semantics while maintaining computational and memory efficiency.

\section{Conclusion}
\label{sec:conclusion}

In this paper, we presented Gen-LangSplat, a generalized framework for efficient, language-grounded 3D Gaussian Splatting. By replacing scene-specific autoencoders with a single pre-trained feature compression module, our method removes the need for per-scene training while maintaining comparable or superior performance to the SOTA, LangSplat, achieving nearly 2× higher overall efficiency and reduced memory cost. Furthermore, extensive ablation studies show that a 16-dimensional latent embedding provides the best trade-off between semantic retention and efficiency, enabling scalable and generalizable open-vocabulary 3D understanding.

{
    \small
    \bibliographystyle{unsrt}
    \bibliography{egbib}
}

\clearpage

\end{document}